%% file: example_paper.tex
\documentclass{article}

\usepackage{microtype}
\usepackage{graphicx}
\usepackage{subcaption}
\usepackage{booktabs} 

\usepackage{hyperref}

\usepackage[preprint]{icml2026}

\usepackage{amsmath}
\usepackage{amssymb}
\usepackage{mathtools}
\usepackage{amsthm}

\usepackage[capitalize,noabbrev]{cleveref}

\theoremstyle{plain}

\theoremstyle{definition}

\theoremstyle{remark}

\usepackage[textsize=tiny]{todonotes}

\icmltitlerunning{~}

\usepackage{booktabs}
\usepackage{multirow}
\usepackage[table,xcdraw]{xcolor}
\usepackage{cuted} 
\usepackage{fontawesome5}

\begin{document}

\twocolumn[
  \icmltitle{RoboArmGS: High-Quality Robotic Arm Splatting via Bézier Curve Refinement}



  \icmlsetsymbol{equal}{*}

  \begin{icmlauthorlist}
    \icmlauthor{Hao Wang}{equal,pek}
    \icmlauthor{Xiaobao Wei}{equal,pek}
    \icmlauthor{Ying Li}{pek}
    \icmlauthor{Qingpo Wuwu}{pek}
    \icmlauthor{Dongli Wu}{pek}
    \icmlauthor{Jiajun Cao}{pek}\\
    \icmlauthor{Ming Lu}{pek}
    \icmlauthor{Wenzhao Zheng}{ucb}
    \icmlauthor{Shanghang Zhang}{pek}
  \end{icmlauthorlist}

  \icmlaffiliation{pek}{Peking University}
  \icmlaffiliation{ucb}{University of California, Berkeley}

  \icmlcorrespondingauthor{Shanghang Zhang}{shanghang@pku.edu.cn}

  \icmlkeywords{Machine Learning, ICML}

  \vskip 0.3in
]



\printAffiliationsAndNotice{}  

\begin{strip}
  \vspace{-20mm}
  \centering
  \includegraphics[width=1.\textwidth ]{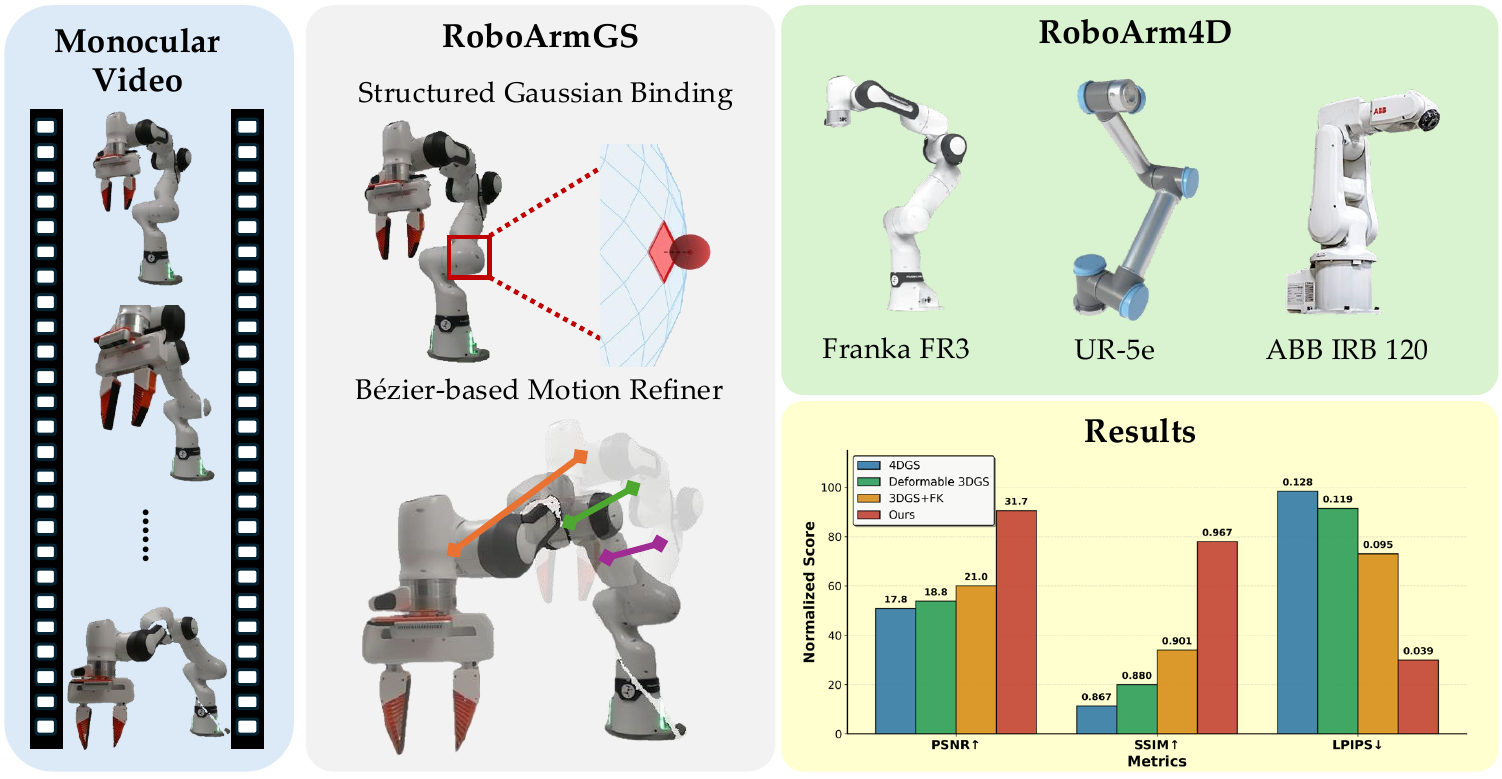}
  \vspace{-5mm}
  \captionsetup{type=figure}
    \captionof{figure}{\textbf{The illustration of RoboArmGS and RoboArm4D.} 
    We refine URDF-rigged motion with learnable Bézier curves that correct per-joint residuals, bridging the gap between idealized kinematics and noisy real-world dynamics. This enables accurate motion modeling and coherent 3D Gaussian binding across arm parts. We evaluate on RoboArm4D, our carefully collected dataset of widely used robotic arms, achieving state-of-the-art rendering quality.}
  \label{fig: teaser}
\end{strip}

\begin{abstract}
Constructing photorealistic and controllable robotic arm digital assets from real observations is fundamental to robotic applications. Current approaches naively bind static 3D Gaussians according to URDF links, forcing them to follow an URDF-rigged motion passively. However, the idealized URDF-rigged motion cannot accurately model the actual motion captured in real-world observations, leading to severe rendering artifacts in 3D Gaussians. To address these challenges, we propose RoboArmGS, a novel hybrid representation that refines the URDF-rigged motion with learnable Bézier curves, enabling more accurate real-world motion modeling. To be more specific, we present a learnable Bézier Curve motion refiner that corrects per-joint residuals to address mismatches between real-world motion and URDF-rigged motion. RoboArmGS enables the learning of more accurate real-world motion while achieving a coherent binding of 3D Gaussians across arm parts. To support future research, we contribute a carefully collected dataset named RoboArm4D, which comprises several widely used robotic arms for evaluating the quality of building high-quality digital assets. We evaluate our approach on RoboArm4D, and RoboArmGS achieves state-of-the-art performance in real-world motion modeling and rendering quality. The code and dataset will be released.
\end{abstract}


\section{Introduction}
\label{sec:intro}

Robotic arms play a crucial role as the primary executors of tasks in contemporary automation and intelligent systems, making them a significant area of focus in the fields of robotics~\cite{liang2024rapid,ma2024hierarchical,sun2025robava,jiang2024robots}. 
Reconstructing high-fidelity, interactable digital assets of robotic arms from real-world observations is a foundational step for creating robust simulation platforms, which is critical for applications in robot control, policy learning, and system monitoring~\cite{wang2025embodiedreamer,pfaff2025scalable,abou2025real}. 
By generating these digital assets from video inputs, we can build simulation environments that faithfully mirror their real-world counterparts. 
This capability is essential for bridging the sim-to-real gap, yet is fundamentally hindered by the discrepancy between idealized kinematic models and physical execution. 
While standard calibration techniques effectively rectify static geometric errors, they often fall short in capturing the complex dynamic deviations inherent in real-world motion. 
Consequently, the challenge of automatically generating dynamic, high-fidelity assets from video remains a key bottleneck in robotics~\cite{xie2025vid2sim, jiang2025phystwin}.

Current approaches for building dynamic digital assets of robotic arms typically involve complex, multi-stage pipelines, which hinder their scalability and robustness~\cite{lou2025robo, han2025re, li2024robogsim}. 
These pipelines usually begin with a strictly controlled data capture phase, requiring multi-view images or videos of the scene.
Subsequently, the robotic arm requires precise segmentation to isolate it from the background and distinguish its individual articulated parts, a process that necessitates extensive labor-intensive manual annotation~\cite{yang2025novel,lou2025robo}.
Such requirements not only limit the method's applicability outside laboratory settings but also introduce significant fragility.
Crucially, this paradigm often assumes a perfect correspondence between the kinematic model and visual observations, thereby failing to account for dynamic discrepancies caused by real-world factors like control latency or joint friction. 

In contrast, learning directly from a single, casually captured monocular video offers a more practical and scalable paradigm~\cite{li2025scalable, tao2025robopearls}. 
While this approach drastically lowers the data collection barrier, it significantly intensifies the challenge of aligning the rigid kinematic prior with the actual, non-ideal motion observed in the footage.
Addressing this requires accurately modeling these dynamic deviations to reconcile them with the standard Universal Robot Description Format (URDF), a step vital for enabling high-fidelity, motion-accurate simulation.

To address these challenges, we propose \textbf{RoboArmGS} (Fig.~\ref{fig: teaser}), a novel hybrid representation designed to harmonize the rigid kinematic prior with actual visual observations.
Our approach dynamically refines the idealized URDF-driven motion using a learnable Bézier curve model, enabling the creation of digital assets that are temporally coherent and precisely aligned with real-world footage.
Central to our approach are two key insights:
(1) Instead of treating 3D Gaussians as an unstructured collection, we introduce a structured binding mechanism that anchors them to a geometric prior, specifically the robot's mesh, to fundamentally enforce topological consistency during motion.
(2) The complex, time-varying discrepancies between the idealized URDF model and the physical robot's dynamics can be explicitly captured and corrected using a flexible, smooth parameterization. 

Guided by these insights, our architecture integrates two core modules.
First, the Structured Gaussian Binding (SGB) anchors each Gaussian to a face on the robot's mesh within its local coordinate frame.
This ensures that the model inherits the robot's rigid geometric structure while retaining the flexibility to capture intricate appearances via learnable offsets.
Second, the Bézier-based Motion Refiner (BMR) addresses dynamic discrepancies by acting as a lightweight motion residual field.
It leverages URDF kinematics as a strong motion prior and learns a smooth, continuous-time residual transformation to reconcile the rigid model with actual observations, thereby compensating for systematic errors in real-world motion.
By synergizing these components, RoboArmGS produces digital assets that are both structurally consistent and kinematically accurate, significantly reducing rendering artifacts caused by motion mismatch.

To validate our approach and support future research, we present \textbf{RoboArm4D}, a specialized dataset featuring representative robotic arms. Designed as a rigorous benchmark for high-fidelity asset creation, it specifically targets the evaluation of how well the generated digital assets conform to real-world visual dynamics.

Our main contributions are as follows:

\begin{itemize}
\item We propose RoboArmGS, which synergizes Structured Gaussian Binding (SGB) and Bézier-based Motion Refiner (BMR) to enforce geometric consistency while correcting dynamic discrepancies, thereby reconciling rigid kinematic priors with actual observations.

\item We contribute RoboArm4D, a benchmark tailored for robotic arm digitization. Featuring representative mainstream robotic arms with distinct morphologies, it evaluates rendering fidelity and motion accuracy against physical reality.

\item Extensive experiments on RoboArm4D demonstrate that RoboArmGS outperforms state-of-the-art methods, setting a new standard for photorealistic and physically aligned simulation.
\end{itemize}


\section{Related Work}

\paragraph{Dynamic Scene Modeling with Gaussian Splatting.} 
3D Gaussian Splatting (3DGS)~\cite{kerbl20233d} has emerged as a dominant representation for high-fidelity scene modeling~\cite{wei2025emd, wei2025gazeGaussian, wang2025plgs}. 
To extend this to dynamic environments, Deformable 3DGS~\cite{yang2024deformable} introduces deformation fields to capture monocular temporal variations, while others~\cite{zhang2024dynamic, wei2025graphavatar} leverage graph neural networks to learn object dynamics under physical interactions.
In the realm of articulated avatars, mesh-guided approaches have shown significant promise. For instance, GaussianAvatars~\cite{qian2024gaussianavatars, chen2025mixedgaussianavatar} bind 3D Gaussians to the FLAME mesh for precise facial control, and Animatable Gaussians~\cite{li2024animatable} apply similar binding strategies to SMPL-based bodies. These methods demonstrate the efficacy of anchoring Gaussians to a geometric prior, a principle we adapt for robotic structures.
In urban scene reconstruction, methods like S$^3$Gaussian~\cite{huang2024s3Gaussian}, StreetGaussians~\cite{yan2024street}, and HUGS~\cite{zhou2024hugs} decompose dynamic entities (e.g., vehicles) from static backgrounds to achieve high-quality rendering and editing. 
More recently, BézierGS~\cite{ma2025b} utilizes learnable Bézier curves to parameterize the global motion trajectories of vehicles, enforcing geometric consistency via inter-group losses. 
While BézierGS employs curves to model the absolute motion path, our approach fundamentally differs by using Bézier curves to model the residual deviation from a kinematic prior. Specifically, we leverage the curves to refine an idealized forward kinematics chain, correcting the dynamic mismatch between the theoretical model and real-world observations.

\vspace{-1em}
\paragraph{Robotic Synthesis using 3D Reconstruction.} 
High-fidelity 3D reconstruction is pivotal for robotic policy learning and data synthesis~\cite{lu2024manigaussian,yu2025manigaussian++,chai2025gaf,wang2025embodiedocc++}.  
Recent approaches have integrated 3DGS to bridge the gap between simulation and reality. 
Several methods focus on integrating 3DGS into existing simulators: SplatSim~\cite{qureshi2025splatsim} replaces mesh rendering with 3DGS for photorealistic output, while Real2Render2Real~\cite{yu2025real2render2real} leverages scanned objects and human demonstrations to scale up training data generation in IsaacLab.
However, for asset creation, most existing pipelines rely on complex multi-view setups. For instance, RoboGSim~\cite{li2024robogsim} and RE³SIM~\cite{han2025re} reconstruct scene assets from multi-view videos to build digital twins. Similarly, RoboSplat~\cite{yang2025novel} uses a unified Gaussian representation from multi-view inputs to model the entire workspace.
While effective, these multi-stage pipelines often suffer from scalability issues.
On the other hand, monocular approaches like ManipDreamer3D~\cite{li2025manipdreamer3d, li2025manipdreamer} synthesize videos from single-view occupancy but focus less on precise kinematic control.
Most relevant to our work is Robo-GS~\cite{lou2025robo}, which binds Gaussians to meshes to reconstruct robotic arms from monocular video. However, Robo-GS primarily targets static asset generation and relies on precise panoramic annotations. 
In contrast, our RoboArmGS not only utilizes monocular input without such heavy supervision but also explicitly models the dynamic discrepancies between the kinematic prior and real-world motion, ensuring both visual and kinematic fidelity during execution.


\section{Methodology}
\begin{figure*}[ht]
  \centering
  \includegraphics[width=1.\textwidth]{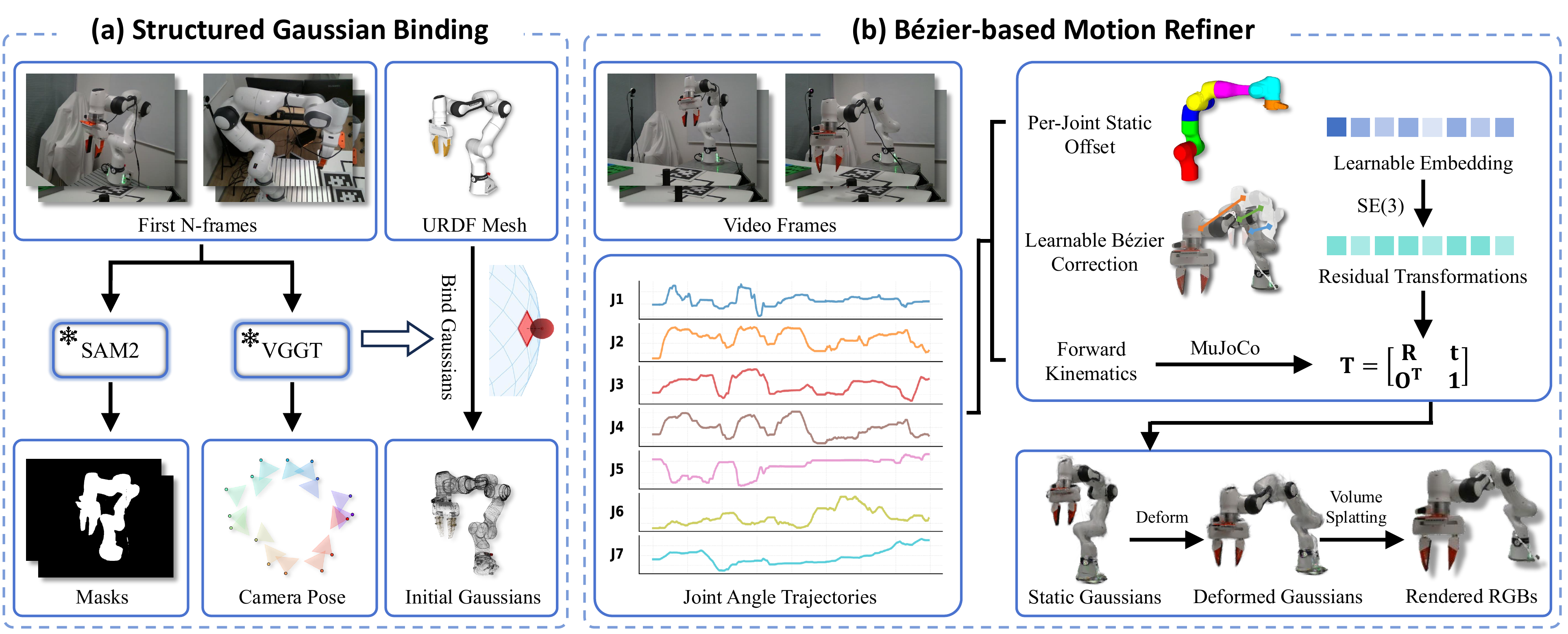}
  \vspace{-1em}
  \caption{\textbf{Overview of RoboArmGS.} (a) SGB anchors Gaussians to URDF links with binding-aware densification for structural consistency. (b) BMR refines kinematic discrepancies via learnable Bézier curves. Jointly optimized from monocular input, our method enables photorealistic rendering for novel robot poses and viewpoints.}
  \label{fig:method}
  \vspace{-1em}
\end{figure*}

In this section, we present RoboArmGS as shown in Fig.~\ref{fig:method}, a framework designed to bridge the sim-to-real motion discrepancy in high-fidelity robotic asset creation.
Instead of relying solely on rigid kinematic assumptions, we introduce a hybrid representation that harmonizes explicit geometric priors with learnable dynamic refinements.
Specifically, Structured Gaussian Binding (SGB) enforces topological consistency by anchoring Gaussians to the robot's mesh, while the Bézier-based Motion Refiner (BMR) explicitly captures continuous residuals between idealized Foward Kinematic (FK) models and real-world observations. 
Finally, we detail the optimization and regularization strategies that ensure robust training from monocular video.

\subsection{Preliminaries}
\paragraph{3D Gaussian Splatting.}
3D Gaussian Splatting~\citep{kerbl20233d} is a real-time radiance field method that represents a scene as a collection of unstructured 3D Gaussians. Each Gaussian is parameterized by a position $\mu$, rotation $q$, anisotropic scaling $s$, opacity $\alpha$, and Spherical Harmonics for view-dependent color. During optimization, a valid positive semi-definite covariance matrix $\Sigma$ is constructed from the learnable scaling and rotation components. To render an image, these 3D Gaussians are projected onto the 2D image plane, sorted by depth, and then composited to compute the final pixel color $C$ via alpha blending:
\begin{equation}
    C = \sum_{i \in N} c_i \alpha'_i \prod_{j=1}^{i-1}(1 - \alpha'_j)
\end{equation}
where $c_i$ is the color of the $i$-th Gaussian and $\alpha'_i$ is its projected opacity. This entire process is made differentiable and efficient by a dedicated tile-based rasterizer.

\paragraph{Bézier Curves.}
A Bézier curve~\citep{mortenson1999mathematics} is a smooth, differentiable parametric curve defined by a set of $n+1$ control points $\{P_i\}_{i=0}^n$. A point on the curve is computed as a weighted sum of these control points using Bernstein basis polynomials as weights:
\begin{equation}
    B(t) = \sum_{i=0}^{n} B_{i,n}(t) P_i, \quad t \in [0, 1]
\end{equation}
where $B_{i,n}(t) = \binom{n}{i} t^i (1-t)^{n-i}$. As $t$ varies from 0 to 1, the curve provides a compact and continuous parameterization that smoothly interpolates within the convex hull of its control points. This makes it an effective tool for modeling temporally coherent trajectories and motion residuals.

\subsection{Structured Gaussian Binding}
Unlike existing methods that naively bind static 3D Gaussians to URDF links, our Structured Gaussian Binding (SGB) is designed to preserve the piecewise rigid structure of robotic arms. SGB explicitly binds each Gaussian to a specific face on the robot's mesh, driven by physics-aware local parameterization and binding-aware adaptive densification. This mesh-level binding ensures all Gaussians on a single link undergo a collective rigid transformation, guaranteeing structural coherence and high-fidelity representation of the articulated motion.

\paragraph{Physics-driven Local Parameterization.}
While inspired by recent advances in rigged avatars~\citep{qian2024gaussianavatars, xu2024gaussian}, we introduce a critical adaptation for rigid-body robotic systems.
Unlike Linear Blend Skinning (LBS)~\citep{li2017learning} models that often introduce non-rigid artifacts, we drive the motion using the robot's forward kinematics to strictly maintain piecewise rigidity.
Specifically, we bind each 3D Gaussian $j$ to a specific face $i$ on the kinematic mesh. This allows us to decompose the Gaussian properties into dynamic, pose-dependent states and learnable, static attributes.
The dynamic properties consist of the local origin $T_i(t)$ and orientation $R_i(t)$, which are determined by the link poses obtained from the MuJoCo engine~\citep{todorov2012mujoco} at time $t$.
The static properties are defined as learnable local offsets that include position $\mu_j$, rotation $r_j$, scale $s_j$, color, and opacity. The final world-space state of a Gaussian for rendering is computed as:
\begin{align}
    \mu'_j(t) &= R_i(t) \mu_j + T_i(t) \label{eq:pos} \\
    r'_j(t) &= R_i(t) r_j \label{eq:rot} \\
    s'_j(t) &= s_j \label{eq:scale}
\end{align}
where $\mu'_j(t)$, $r'_j(t)$, and $s'_j(t)$ denote the world-space parameters. This ensures that all Gaussians anchored to the same link inherit a collective rigid transformation, while the learnable offsets $\mu_j, r_j$ provide the degrees of freedom necessary to reconstruct complex surface geometries that exceed the resolution of the proxy URDF mesh.

\paragraph{Binding-Aware Adaptive Densification.} 
To capture high-frequency surface details while respecting the robot's rigid structure, we adapt the density control mechanism of 3DGS~\citep{kerbl20233d} through two binding-aware modifications. 
First, we implement binding inheritance, ensuring that newly densified primitives automatically inherit the face index of their progenitor. This anchoring mechanism preserves piecewise rigid motion by preventing Gaussians from drifting away from their designated kinematic links during splitting or cloning. 
Second, we introduce a structural preservation constraint that prevents the pruning of the final Gaussian associated with any mesh face. This safeguard is critical for maintaining the geometric integrity of the robotic structure, effectively eliminating visual "holes" that might otherwise emerge during large-scale articulated movements. 
Together, these adaptations ensure that the optimization process yields a high-fidelity representation that strictly adheres to the robot's physical configuration and topological constraints.

\subsection{Bézier-based Motion Refiner}
To reconcile idealized kinematic models with physical reality, we introduce the Bézier-based Motion Refiner (BMR). 
While standard FK provides a strong motion prior, they often fail to account for real-world deviations such as control latency or joint friction. 
Drawing inspiration from the smooth temporal parameterization in BézierGS~\cite{ma2025b}, our BMR models these discrepancies as a continuous residual field. 
We employ a hierarchical decomposition that couples a learnable, time-varying global correction with static, per-joint offsets. This design leverages the inherent smoothness of Bézier curves to ensure global temporal coherence while providing the local flexibility necessary to compensate for systematic kinematic errors.

\paragraph{Learnable Bézier Correction.}
The time-varying global correction, $\mathbf{T}_{\text{Bézier}}(t)$, is modeled by a learnable Bézier curve that operates in a 9-dimensional parameter space. Unlike BézierGS~\cite{ma2025b}, which uses Bézier curves to represent the full motion trajectory of objects, we employ the curve here to model the residual SE(3) deviation from the rigid FK prior. We parameterize the residual transformations as 9D vectors $\boldsymbol{\delta} = [\Delta \mathbf{x}, \Delta \mathbf{r}]$, consisting of a 3D translation $\Delta \mathbf{x}$ and a continuous 6D rotation representation $\Delta \mathbf{r}$~\citep{zhou2019continuity}. The Bézier curve, defined by $K+1$ learnable control points $\{\mathbf{p}_k\}_{k=0}^K \subset \mathbb{R}^9$, outputs a 9D residual vector for any time $t$:
\begin{equation}
    \boldsymbol{\delta}_{\text{Bézier}}(t) = \omega \sum_{k=0}^{K} B_k^K(t) \cdot \mathbf{p}_k
\end{equation}
where $B_k^K(t)$ are the Bernstein basis polynomials. We use a high-order curve ($K=19$) to capture complex motion patterns and scale the output by an influence factor $\omega$ to ensure it acts as a refinement. The resulting 9D vector is then converted to a valid SE(3) matrix $\mathbf{T}_{\text{Bézier}}(t)$ via Gram-Schmidt orthogonalization for the rotational part.

\paragraph{Per-Joint Static Correction.}
To address the articulated structure of the robot, we introduce a static, per-joint offset $\mathbf{T}_{\text{embed}}^{(k)}$, represented by a learnable embedding $\mathbf{e}^{(k)} \in \mathbb{R}^9$ for each moving joint $k$. 
These time-invariant embeddings are optimized to correct for intrinsic structural discrepancies, such as link length errors or joint calibration drift. 
By integrating these local offsets into the kinematic chain alongside the global Bézier correction, our model specifically accounts for the hierarchical dependencies of robotic motion—a feature that distinguishes our approach from general, unstructured trajectory modeling.

\paragraph{Final Pose Composition.}
The final refined pose for each joint is obtained by integrating the nominal FK pose from MuJoCo with our two hierarchical corrections. 
The composition order is designed to handle errors at different levels of the kinematic chain: the global Bézier correction first compensates for temporal drifts or base misalignments, while the static embedding further refines the local joint-to-link transformation. For a specific joint $k$, the final world-space transformation is formulated as
\begin{equation}
    \mathbf{T}_{\text{final}}^{(k)}(t) = \mathbf{T}_{\text{FK}}^{(g)}(t) \circ \mathbf{T}_{\text{Bézier}}(t) \circ \mathbf{T}_{\text{embed}}^{(k)} \label{eq:final_pose_composition}
\end{equation}
where $\mathbf{T}_{\text{FK}}^{(g)}(t)$ denotes the global pose derived from the URDF model. This unified formulation ensures that the digital asset remains kinematically sound while precisely aligning with the visual observations in the video.

\subsection{Optimization and Regularization}
The overall optimization of our model is supervised by a combination of a primary rendering loss and several crucial regularization terms designed to ensure geometric stability and temporal coherence.

\paragraph{Rendering Loss.}
The primary supervision signal is a rendering loss comparing the rendered images with ground truth frames. Following standard practice~\citep{kerbl20233d}, we use a combination of an L1 photometric loss and a D-SSIM term:
\begin{equation}
    \mathcal{L}_{\text{rgb}} = (1 - \lambda) \mathcal{L}_1 + \lambda \mathcal{L}_{\text{D-SSIM}} \quad (\lambda=0.2)
\end{equation}

\paragraph{Geometric Regularization.}
To enforce the topological constraints of SGB module and ensure Gaussians remain well-aligned with their parent mesh faces, we adopt two regularization terms inspired by~\citep{qian2024gaussianavatars}. We penalize the local position offset $\mu_j$ and scale $s_j$ if they exceed predefined thresholds, preventing primitives from detaching from the rigid structure or exhibiting visual jitter:
\begin{align}
    \mathcal{L}_{\text{pos}} &= \| \max(|\mu_j| - \epsilon_{\text{pos}}, 0) \|^2_2 \\
    \mathcal{L}_{\text{scale}} &= \| \max(|s_j| - \epsilon_{\text{scale}}, 0) \|^2_2
\end{align}

\paragraph{Bézier Velocity Regularization.}
To enforce that the corrective motions learned by our BMR module are temporally smooth, we regularize the squared L2 norm of the learnable Bézier curve's instantaneous velocity, $\mathbf{v}(t)$. We approximate this velocity using a numerically stable central difference scheme, and define the loss as:
\begin{equation}
    \mathcal{L}_{\text{vel}} = \| \mathbf{v}(t) \|^2_2
\end{equation}
This loss penalizes abrupt changes in the learned 9D residual parameters, encouraging a smooth motion trajectory.

\paragraph{Total Objective.}
Our final training objective is a weighted sum of the aforementioned losses:
\begin{equation}
    \mathcal{L}_{\text{total}} = \mathcal{L}_{\text{rgb}} + \lambda_{\text{pos}} \mathcal{L}_{\text{pos}} + \lambda_{\text{scale}} \mathcal{L}_{\text{scale}} + \lambda_{\text{vel}} \mathcal{L}_{\text{vel}}
\end{equation}
All learnable parameters, including Gaussian attributes, Bézier control points, and per-joint embeddings, are jointly optimized by minimizing this objective.


\section{Experiments}

\input{tables/main_result}

\subsection{Setup}
\paragraph{Settings.}
We evaluate our method, RoboArmGS, across two challenging settings to assess its reconstruction and motion modeling capabilities. 
The first setting is \textbf{Novel-View Synthesis}, where the model is trained on multi-view images of a stationary arm and is evaluated on its ability to render photorealistic images from unseen camera viewpoints. 
The second, more demanding setting is \textbf{Novel-Pose Synthesis}. 
In this scenario, the model is trained on a video sequence of the arm in motion and is then required to render it from a fixed camera viewpoint, but for a sequence of held-out, unseen joint configurations. 
Due to space limitations, please refer to the appendix for more experiments.

\paragraph{Datasets.}
Existing public robotics datasets often lack the precise alignment between visual observations and kinematic states required for high-fidelity asset creation. To facilitate research in this area, we present RoboArm4D, a benchmark designed for the generation and evaluation of dynamic digital assets. This dataset features several widely-used industrial arms, including the Franka Research 3, UR-5e, and ABB IRB 120. For each platform, we provide synchronized data packages comprising monocular video sequences, calibrated camera parameters, time-stamped joint trajectories, and the corresponding URDF files. This collection is intended to support the development of methods that reconcile rigid kinematic priors with actual physical motion. Further details regarding the data collection and package structure are available in the appendix.

\paragraph{Metrics.}
To quantitatively evaluate rendering and motion fidelity, we adopt three standard image-based quality metrics: Peak Signal-to-Noise Ratio (PSNR), Structural Similarity Index Measure (SSIM), and Learned Perceptual Image Patch Similarity (LPIPS). These metrics jointly assess both pixel-level accuracy and perceptual quality of rendered views, providing a comprehensive evaluation of real-world motion consistency and rendering fidelity across static and dynamic robotic scenarios.

\paragraph{Implementation Details.}
Our model, RoboArmGS, is implemented in PyTorch and trained for 600,000 iterations on a single NVIDIA H100 GPU using the Adam optimizer. For 3D Gaussian attributes, we follow the standard learning rate schedule from 3DGS~\citep{kerbl20233d}, with position learning rates exponentially decayed to 1\% of initial values. The BMR parameters are optimized with a learning rate of 0.0015, weight decay of 0.0001, and influence factor $\omega=0.1$. Gaussians are initialized by uniform sampling on the kinematic mesh surface, with binding-aware adaptive densification activated every 100 iterations from iteration 500 to 60,000; unstable Gaussian opacities are reset every 3,000 iterations to prune floaters. FK is computed via MuJoCo~\citep{todorov2012mujoco} engine, and regularization hyperparameters are set as $\lambda_{\text{pos}}=0.01, \epsilon_{\text{pos}}=1.0$, $\lambda_{\text{scale}}=1.0, \epsilon_{\text{scale}}=0.6$, and $\lambda_{\text{vel}}=0.001$.

\begin{figure*}[!ht]
  \centering
  \includegraphics[width=1.\textwidth]{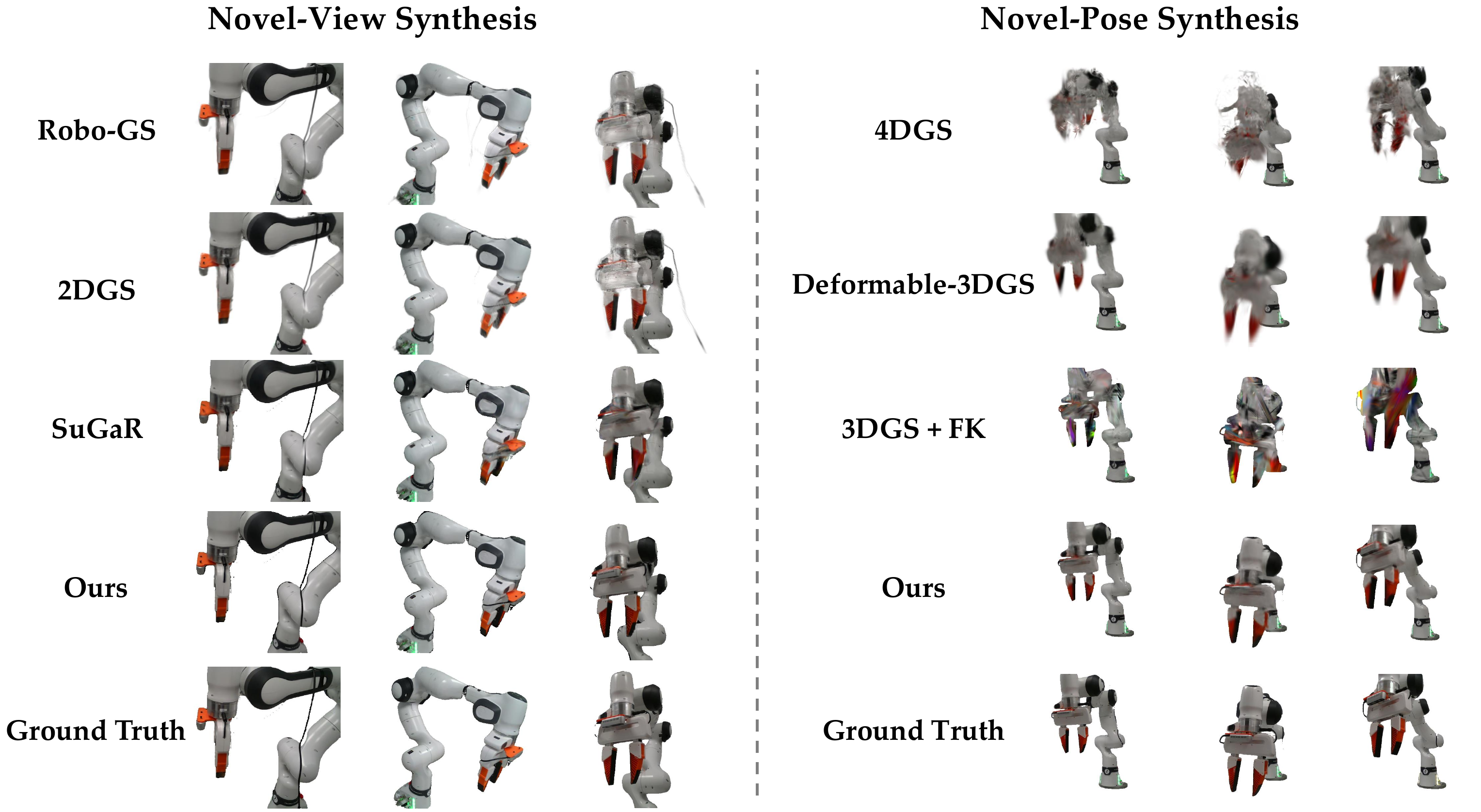}
  \caption{\textbf{Qualitative comparison on novel-view and novel-pose synthesis.} RoboArmGS achieves photorealistic rendering with precise geometric alignment across both tasks. While baseline methods suffer from severe blurring, structural distortion, or rendering artifacts under unseen configurations, our approach maintains high fidelity and kinematic accuracy, ensuring sharp and coherent results.}
  \vspace{-1em}
  \label{fig:vis}
\end{figure*}

\subsection{Main Results}
\paragraph{Novel-View Synthesis.}

We first evaluate static 3D reconstruction fidelity through the novel-view synthesis task. As reported in Tab.~\ref{tab:main_results}, RoboArmGS achieves state-of-the-art results across all metrics and significantly outperforms other methods. Although Robo-GS~\citep{lou2025robo} is a robot-specific method, it still relies on the vanilla 3D Gaussian model for rendering, which performs unconstrained spatial optimization. In contrast, our SGB module explicitly anchors Gaussians to the mesh surface to incorporate the robot's kinematic structure as a strong geometric prior. This structured approach prevents the formation of floating artifacts and maintains strict structural coherence, providing a more robust foundation for dynamic modeling. Qualitative comparisons in Fig.~\ref{fig:vis} further demonstrate that our method reconstructs fine-grained surface details with higher visual fidelity than these unconstrained baselines.

\paragraph{Novel-Pose Synthesis.}
We evaluate dynamic modeling capability through the novel-pose synthesis task, which involves rendering the robot in unseen joint configurations. As shown in Tab.~\ref{tab:main_results}, RoboArmGS substantially outperforms all baselines across all metrics. Generic 4D reconstruction methods~\citep{wu20244d, yang2024deformable} model temporal evolution using time indices rather than kinematic parameters. Consequently, they lack the ability to render from specified joint angles, a capability that is essential for digital assets applications. Even on held-out test frames, these generic deformation models fail to capture complex articulated motions and produce significant artifacts.
To evaluate kinematic-driven approaches, we implement a 3DGS + FK baseline. This baseline binds 3D Gaussians to the mesh and drives their motion via nominal forward kinematics without any motion refinement. While Robo-GS~\citep{lou2025robo} shares a similar philosophy, its reliance on manual part segmentation and intensive parameter tuning limits its scalability. Therefore, we utilize 3DGS + FK as the representative kinematic baseline. Although this approach maintains the robot's rigid structure, it produces significant rendering errors, particularly in color and texture fidelity. These discrepancies arise because the nominal URDF poses deviate from the physical reality in the video, causing the rendered Gaussians to misalign with the actual image content. Our Bézier-based Motion Refiner addresses this by integrating the refinement of robotic motion directly into the 3D Gaussian optimization loop. By jointly optimizing the Bézier parameters alongside Gaussian attributes, our model reconciles nominal kinematics with actual visual dynamics through temporally smooth corrections. As demonstrated in Fig.~\ref{fig:vis}, RoboArmGS produces sharp, structurally coherent images that align precisely with the ground truth, validating its accuracy as a controllable digital asset.

\subsection{Ablation Study}

\begin{figure}[!ht]
  \centering
  \includegraphics[width=0.45\textwidth]{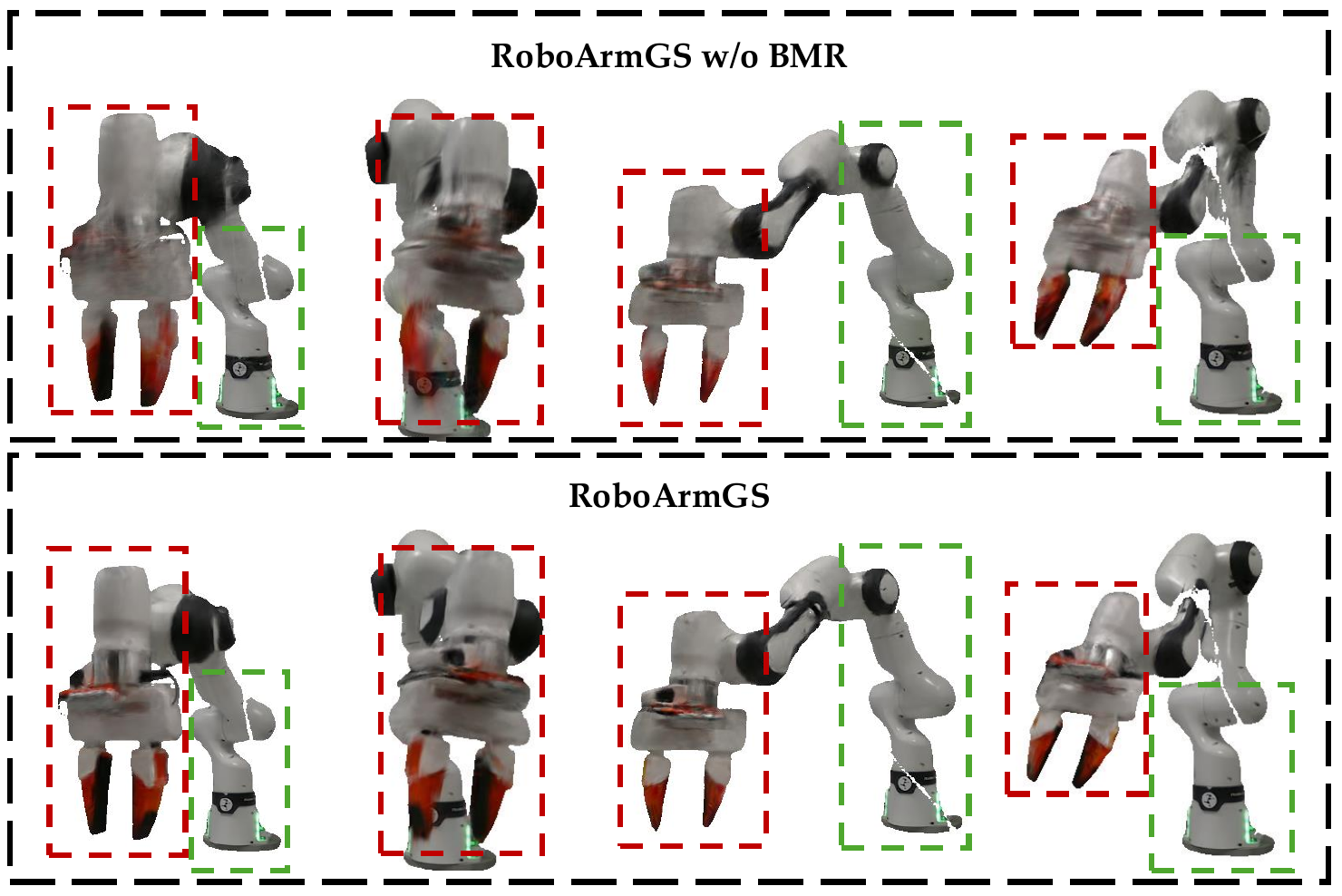}
  \caption{\textbf{Ablation study visualization.} Low-motion regions (green) show consistent results, whereas highly articulated components (red) exhibit artifacts without BMR, underscoring its necessity for complex motion alignment.}
  \label{fig:abla}
  \vspace{-1em}
\end{figure}

\paragraph{Module Effectiveness Analysis. }
\input{tables/ablation}
We validate the contribution of each proposed component through comprehensive ablation studies, as presented in Tab. \ref{tab:ablation_side_by_side_v2}. For the novel-view synthesis task, the removal of the SGB module leads to a severe degradation in reconstruction quality. Further ablating its sub-components, such as adaptive densification and binding mechanisms, confirms their necessity for achieving high-fidelity static representation. Regarding the novel-pose synthesis task, the BMR module plays a critical role in reconciling kinematic discrepancies. As illustrated in Fig. \ref{fig:abla}, the impact of BMR is highly motion-dependent. In regions with minimal movement, such as the robot base (green box), the differences between the full model and the ablated variant are negligible. Conversely, in regions experiencing large-scale articulation, such as the gripper (red box), the absence of BMR results in significant visual artifacts and misalignments. These results demonstrate that the learnable Bézier correction for global dynamics and the per-joint static offsets for local errors are complementary. This study empirically validates that all components of SGB and BMR are indispensable for the photorealistic and kinematically accurate performance of RoboArmGS.

\input{tables/abla_bezier_factor}

\paragraph{Bézier Influence Factor Analysis.}
We examine the sensitivity of the influence factor $\omega$, which modulates the learnable Bézier correction. The results for the novel-pose synthesis task are summarized in Tab. \ref{tab:ablation_alpha}. Setting $\omega=0.0$ effectively disables the global motion refinement, which yields the lowest performance and confirms that nominal kinematics alone are insufficient for precise physical alignment. Even a minimal correction factor of $\omega=0.01$ leads to a substantial improvement across all metrics. The performance peaks at $\omega=0.1$, providing the optimal degree of refinement for aligning the digital asset with visual observations. However, further increasing $\omega$ to $1.0$ results in a slight performance decline, suggesting that an excessively large correction magnitude may introduce training instability or overfitting to specific motion trajectories. Consequently, we select $\omega=0.1$ as the standard setting to ensure a robust balance between motion accuracy and stability.

\begin{figure}[!ht]
  \centering
  \includegraphics[width=0.5\textwidth]{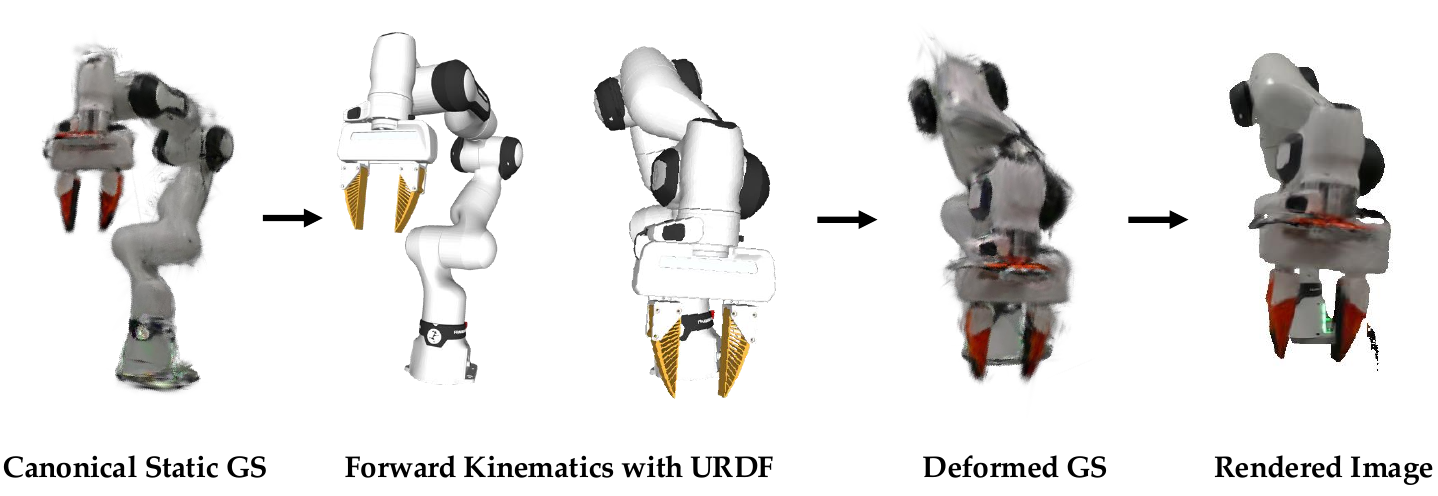}
  \caption{\textbf{Digital Asset Testing Pipeline.} Our trained digital assets enable controllable rendering at test time. Given target joint angles, the URDF-rigged mesh deforms accordingly (left to middle), driving coherent Gaussian transformation via our learned Bézier refinement. The deformed Gaussians render photorealistic images (right) with precise kinematic control.}
  \vspace{-2em}
  \label{fig:asset_vis}
\end{figure}

\subsection{Digital Assets Visualization}
We visualize the interactive testing pipeline of our digital assets in Fig.~\ref{fig:asset_vis}. At test time, users can control the robotic arm in real-time by specifying arbitrary joint angles. The underlying URDF-rigged mesh deforms according to the input kinematics, which are then dynamically refined by our learned Bézier curves to match real-world motion. This refined mesh deformation drives the coherent transformation of all bound 3D Gaussians. These Gaussians are subsequently rendered into photorealistic images from any novel viewpoint, faithfully reflecting the specified pose. By seamlessly integrating a controllable kinematic structure with a high-fidelity neural representation, our digital assets achieve both precise kinematic control and superior rendering quality, enabling flexible and realistic simulation for various downstream applications.


\section{Conclusion}
In this paper, we present RoboArmGS, a hybrid representation that synergizes Structured Gaussian Binding (SGB) with a Bézier-based Motion Refiner (BMR) to achieve both geometric consistency and precise motion modeling. By anchoring Gaussians to the robot's kinematic structure and explicitly modeling physical motion residuals, our method effectively reconciles the discrepancy between nominal URDF models and real-world observations. To support future research, we contribute RoboArm4D, a specialized benchmark designed for the reconstruction and evaluation of dynamic robotic digital assets. Extensive experiments demonstrate that RoboArmGS achieves state-of-the-art performance in both novel-view and novel-pose synthesis, providing a robust and scalable solution for building photorealistic, kinematically accurate digital assets.


\nocite{langley00}

\bibliography{example_paper}
\bibliographystyle{icml2026}

\newpage
\appendix
\onecolumn

\section{Overview}
This appendix provides supplementary details regarding the proposed RoboArm4D dataset and presents additional experimental evaluations for RoboArmGS. We first elaborate on the dataset construction in the following section, covering the hardware specifications, data capture protocols, and the processing pipeline used to generate the monocular sequences. Subsequently, we report extended quantitative and qualitative results on the Universal Robots UR5e and ABB IRB 120 sequences. These additional experiments on Novel-View Synthesis and Novel-Pose Synthesis further validate the generalizability and high-fidelity rendering capabilities of our method across diverse robotic morphologies.

\section{The RoboArm4D Dataset Details}
To facilitate research in high-fidelity robotic arm reconstruction and motion modeling, we introduce the RoboArm4D dataset. This dataset features monocular video sequences of several common robotic arms performing diverse motions. This section provides a detailed overview of our data capture hardware, protocol, and processing pipeline.

\subsection{Hardware Setup}
Our data collection setup was designed for simplicity and accessibility, requiring minimal specialized equipment. We captured all sequences using a single, handheld Intel RealSense L515 camera, recording RGB video at a resolution of 640x480 and a rate of 30 frames per second to simulate casual capture conditions. The dataset features three widely-used robotic arms: the Franka Research 3 (7-DoF), the Universal Robots UR5e (6-DoF), and the ABB IRB 120 (6-DoF). Each arm was mounted on a workbench within a standard laboratory environment, characterized by diffuse overhead lighting and a relatively static, yet typical, background containing various lab equipment.

\subsection{Data Capture Protocol}
\subsubsection{Camera Pose Calibration}
Accurately determining the pose of a static camera is crucial. To achieve this, we first captured a short calibration video (approx. 30 seconds) where the handheld camera was moved extensively around the static robotic arm and its environment. This multi-view sequence was processed using VGGT~\citep{wang2025vggt} to generate a sparse 3D reconstruction of the scene and to precisely compute the camera poses for each frame of the calibration video. From this set of calibrated poses, we selected a single, fixed viewpoint for the subsequent motion capture. This process effectively pre-calibrates the static camera's extrinsic parameters within the scene's coordinate frame.

\subsubsection{Motion Trajectory and Recording}
With the camera now fixed in its pre-calibrated position, the robotic arm was programmed to execute a pre-defined, smooth trajectory. These trajectories were designed to cover a wide range of joint configurations, including both simple single-joint movements and complex multi-joint coordinated motions. While the arm was in motion, we recorded a continuous video sequence from the static viewpoint. Each motion sequence lasts approximately 30-60 seconds, resulting in 900-1800 frames. Simultaneously, joint angles were recorded directly from the robot's controller API to be synchronized with the video frames during post-processing.

\subsubsection{Data Processing Pipeline}
The raw video and joint angle data were processed into a format suitable for training our model using the following steps:
\begin{itemize}
    \item Video to Frames: The captured videos were decomposed into individual PNG frames.
    \item Camera Pose Estimation: We used VGGT~\citep{wang2025vggt} to estimate the camera intrinsics and the per-frame extrinsic poses (camera-to-world transformation). Poses were optimized over the entire sequence to ensure global consistency.
    \item Foreground Segmentation: To separate the robotic arm from the background, we employed the Segment Anything Model 2 (SAM2)~\citep{ravi2024sam}. We provided a few initial keyframe masks, and SAM2~\citep{ravi2024sam} automatically propagated the segmentation to the entire sequence, followed by minor manual refinement where necessary. This resulted in a pixel-perfect foreground mask for each frame.
    \item Data Synchronization: The high-frequency joint angle data were synchronized with the video frames. We used linear interpolation to obtain the precise joint angle configuration corresponding to the capture time of each frame.
\end{itemize}

\subsection{Dataset Splitting Protocol}
We adopt a uniform sampling strategy to partition the dataset into training, validation, and test sets with an 8:1:1 ratio. Specifically, we select every 10th frame for the test set and every 10th frame for the validation set, ensuring distinct frames for each subset (non-overlapping). The remaining frames constitute the training set. This interval-based splitting ensures that the evaluation covers the full range of motion and viewpoints present in the recorded sequences.

\begin{figure*}[!b]
  \centering
  \includegraphics[width=0.9\textwidth]{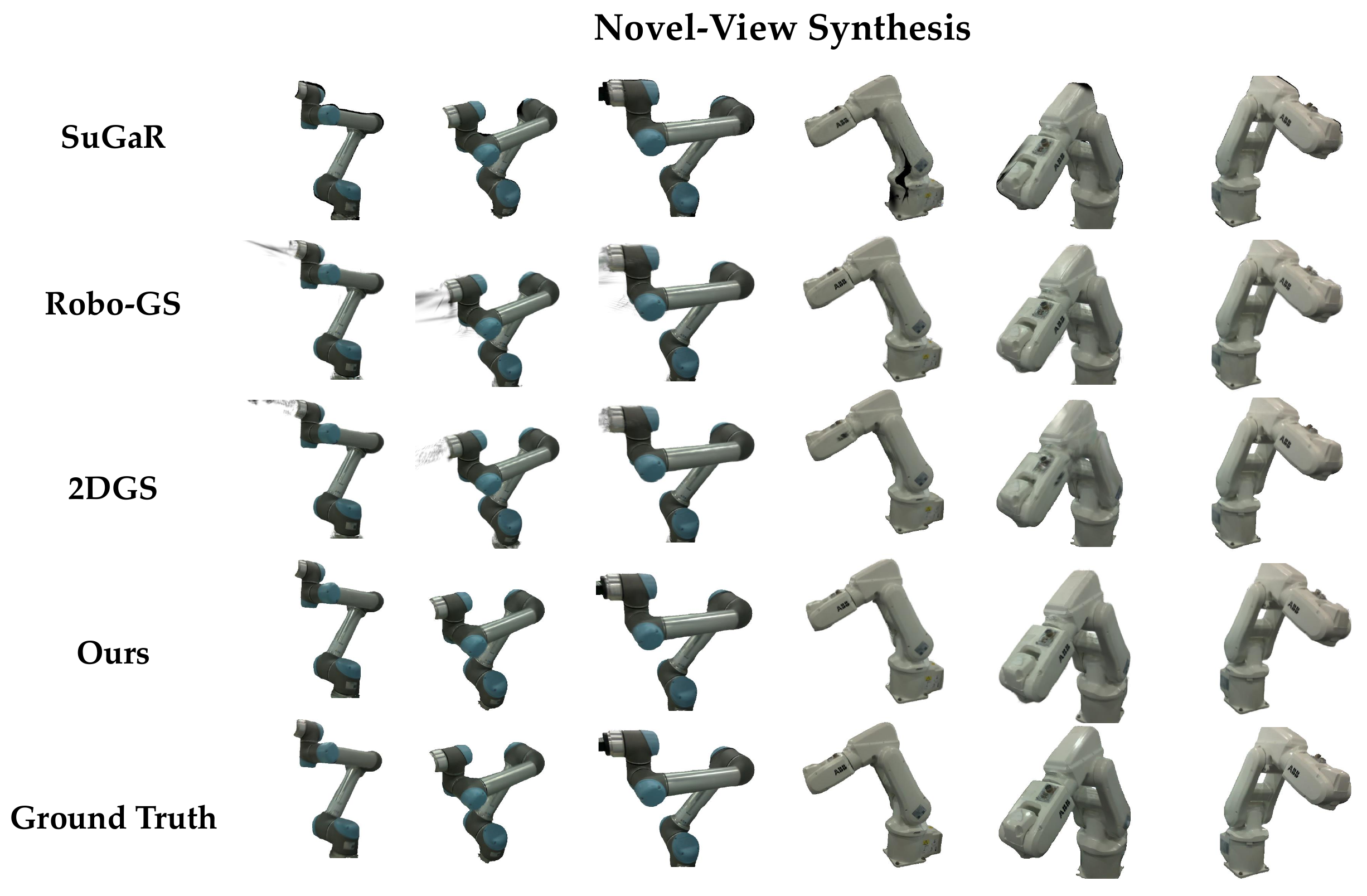}
  \caption{\textbf{Qualitative results for Novel-View Synthesis.} RoboArmGS synthesizes photorealistic images from unseen viewpoints for both the Universal Robots UR5e (left) and ABB IRB 120 (right). Our method faithfully reconstructs high-frequency details and preserves the visual fidelity of the robot's appearance, while maintaining sharp geometric boundaries against the background.}
  \label{fig:app_nvs}
\end{figure*}

\section{Additional Experiments Results}
In this section, we provide a detailed per-scene quantitative analysis of RoboArmGS on the RoboArm4D dataset, specifically focusing on the Universal Robots UR5e and ABB IRB 120 sequences. As shown in Tab.~\ref{tab:ur5e_abb_comparison}, our method achieves consistently superior performance across different robotic morphologies, significantly outperforming existing state-of-the-art (SOTA) methods and validating the generalizability of our proposed framework.

\input{tables/app_results}

\begin{figure*}[!ht]
  \centering
  \includegraphics[width=0.9\textwidth]{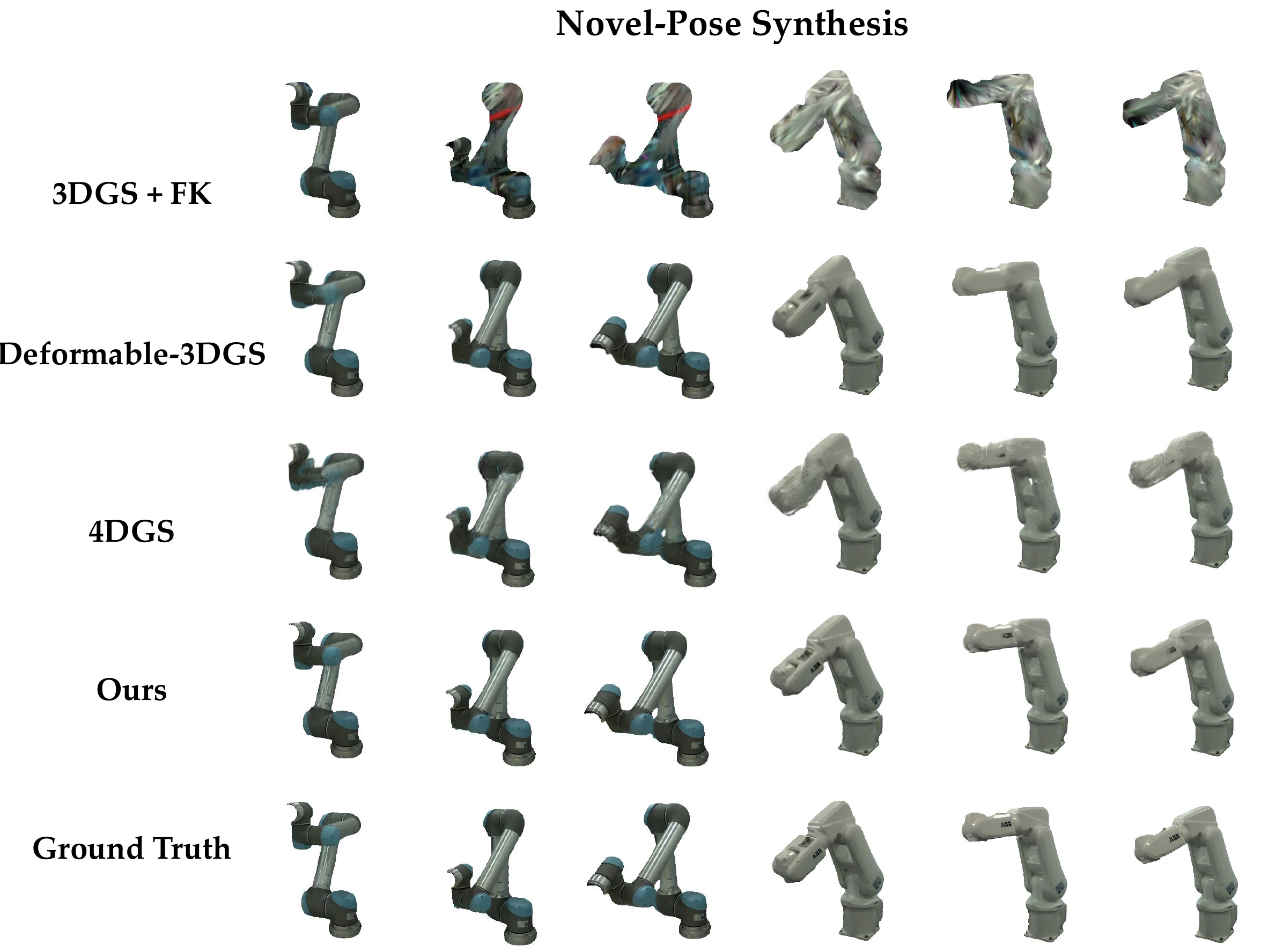}
  \caption{\textbf{Qualitative results for Novel-Pose Synthesis.} We visualize the rendered results of the robotic arms under unseen joint configurations (held-out test poses). Even in challenging poses that deviate significantly from the canonical training configurations, RoboArmGS maintains structural integrity and visual fidelity. The results demonstrate precise kinematic alignment and realistic shading changes consistent with the robot's motion.}
  \label{fig:app_nps}
\end{figure*}

\subsection{Novel-View Synthesis}
We first evaluate the static reconstruction quality on the Novel-View Synthesis task. As reported in Tab.~\ref{tab:ur5e_abb_comparison} (left), RoboArmGS achieves leading rendering fidelity on both robotic arms. Specifically, for the UR5e, our method attains a PSNR of 35.220 dB, surpassing the best baseline (SuGaR) by a substantial margin of 4.768 dB. Similarly, for the ABB IRB 120, we achieve a PSNR of 34.815 dB, outperforming Robo-GS by 4.628 dB.
These metrics indicate that our Structured Gaussian Binding (SGB) strategy effectively leverages the underlying kinematic mesh to provide strong geometric constraints for the 3D Gaussians. Unlike unconstrained methods like 2DGS or SuGaR, SGB allows for precise geometry reconstruction even from sparse monocular inputs. Qualitative results (see Fig.~\ref{fig:app_nvs}) further corroborate these findings: our method renders sharp boundaries and preserves fine-grained texture details, effectively capturing the specific material characteristics of each robot without the "cloudy" floaters or structural artifacts often seen in previous Gaussian-based reconstructions.

\subsection{Novel-Pose Synthesis}
The Novel-Pose Synthesis task evaluates the model's ability to generate photorealistic images under unseen joint configurations, which is critical for high-fidelity digital assets simulation.
Remarkably, as shown in Tab.~\ref{tab:ur5e_abb_comparison} (right), RoboArmGS achieves even more better results on this dynamic task. For the UR5e arm, our method achieves a PSNR of 39.559 dB, which is a massive 8.6 dB improvement over the second-best method (Deformable 3DGS, 30.959 dB). On the ABB arm, we also maintain a significant lead with 36.379 dB PSNR. The LPIPS scores for both scenes remain exceptionally low, reflecting near-perfect perceptual realism.
This superior performance is primarily attributed to our Bézier-based Motion Refiner (BMR). While methods like 4DGS and Deformable 3DGS struggle with large-range articulated movements and cumulative joint errors, BMR successfully compensates for the residual offsets between the idealized URDF model and real-world mechanical motion. By decoupling these motion artifacts from appearance learning, BMR ensures that Gaussians remain accurately attached to their respective links during complex movements. As visualized in Fig.~\ref{fig:app_nps}, our rendered arms align perfectly with the target poses while maintaining consistent lighting and occlusion handling, proving that RoboArmGS is a highly robust solution for dynamic robotic simulation.


\end{document}

%% file: tables/main_result.tex
\begin{table*}[t]
\centering
\resizebox{\textwidth}{!}{
\begin{tabular}{l|ccc|c|l|ccc}
\toprule
\multicolumn{4}{c|}{\textbf{Novel-View Synthesis}} & \phantom{a} & \multicolumn{4}{c}{\textbf{Novel-Pose Synthesis}} \\
\cmidrule(r){1-4} \cmidrule(l){6-9}
\textbf{Method} & PSNR↑ & SSIM↑ & LPIPS↓ & & \textbf{Method} & PSNR↑ & SSIM↑ & LPIPS↓ \\
\midrule
Robo-GS~\citep{lou2025robo} & 26.522 & 0.918 & 0.099 && 4DGS~\citep{wu20244d} & 17.812 & 0.867 & 0.128 \\
2DGS~\citep{huang20242d} & 26.798 & 0.920 & 0.101 && Deformable 3DGS~\citep{yang2024deformable} & 18.844 & 0.880 & 0.119 \\
SuGaR~\citep{guedon2024sugar} & \cellcolor[HTML]{FFF2CC}26.969 & \cellcolor[HTML]{FFF2CC}0.929 & \cellcolor[HTML]{FFF2CC}0.085 && 3DGS + FK & \cellcolor[HTML]{FFF2CC}21.038 & \cellcolor[HTML]{FFF2CC}0.901 & \cellcolor[HTML]{FFF2CC}0.095 \\
\midrule
Ours & \cellcolor[HTML]{E2EFDA}\textbf{28.669} & \cellcolor[HTML]{E2EFDA}\textbf{0.938} & \cellcolor[HTML]{E2EFDA}\textbf{0.065} && Ours & \cellcolor[HTML]{E2EFDA}\textbf{31.704} & \cellcolor[HTML]{E2EFDA}\textbf{0.967} & \cellcolor[HTML]{E2EFDA}\textbf{0.039} \\
\bottomrule
\end{tabular}%
} 
\caption{\textbf{Quantitative comparison.} Our model demonstrates superior performance in both static Novel-View Synthesis (left) and dynamic Novel-Pose Synthesis (right). \colorbox[HTML]{E2EFDA}{Green} indicates the best and \colorbox[HTML]{FFF2CC}{yellow} indicates the second-best performance.}
\vspace{-2em}
\label{tab:main_results}
\end{table*}

%% file: tables/ablation.tex
\begin{table*}[t]
\centering
\small
\renewcommand{\arraystretch}{1.1}
\setlength{\tabcolsep}{6pt} 
\begin{tabular}{l|ccc|c|l|ccc}
\toprule
\multicolumn{4}{c|}{\textbf{Novel-View Synthesis Ablation}} & \phantom{a} & \multicolumn{4}{c}{\textbf{Novel-Pose Synthesis Ablation}} \\
\cmidrule(r){1-4} \cmidrule(l){6-9}
\textbf{Model Variant} & PSNR↑ & SSIM↑ & LPIPS↓ & & \textbf{Model Variant} & PSNR↑ & SSIM↑ & LPIPS↓ \\

\midrule
w/o SGB  & 20.452 & 0.809 & 0.438 && w/o BMR  & \cellcolor[HTML]{FFF2CC}29.565 &\cellcolor[HTML]{FFF2CC}0.957 & 0.052 \\
\hspace{1em} - w/o Adaptive Densification & 20.414 & 0.803 & 0.451 && \hspace{1em} - w/o Bézier Correction & 29.384 & 0.956 & 0.052 \\
\hspace{1em} - w/o Binding & \cellcolor[HTML]{FFF2CC}21.621 & \cellcolor[HTML]{FFF2CC}0.831 & \cellcolor[HTML]{FFF2CC}0.205 && \hspace{1em} - w/o Per-Joint Offset & 29.418 & 0.956 & \cellcolor[HTML]{FFF2CC}0.053 \\
\midrule
Ours (Full Model) & \cellcolor[HTML]{E2EFDA}\textbf{28.669} & \cellcolor[HTML]{E2EFDA}\textbf{0.938} & \cellcolor[HTML]{E2EFDA}\textbf{0.065} && Ours (Full Model) & \cellcolor[HTML]{E2EFDA}\textbf{31.704} & \cellcolor[HTML]{E2EFDA}\textbf{0.967} & \cellcolor[HTML]{E2EFDA}\textbf{0.039} \\
\bottomrule
\end{tabular}
\caption{\textbf{Quantitative comparison of module effectiveness analysis.} We validate the contributions of our key components on both tasks. The SGB module and its sub-components are crucial for high-quality static reconstruction. The BMR and its sub-components are essential for correcting kinematic errors in novel-pose synthesis. \colorbox[HTML]{E2EFDA}{Green} indicates the best and \colorbox[HTML]{FFF2CC}{yellow} indicates the second-best performance.}
\vspace{-2em}
\label{tab:ablation_side_by_side_v2}
\end{table*}

%% file: tables/abla_bezier_factor.tex
\begin{table}[!ht]
\centering
\small
\renewcommand{\arraystretch}{1.1}
\setlength{\tabcolsep}{10pt}
\begin{tabular}{c|ccc}
\toprule
\textbf{Influence Factor ($\omega$)} & PSNR↑ & SSIM↑ & LPIPS↓ \\
\midrule
$\omega = 0.0$ & 29.384 & 0.956 & 0.052 \\
$\omega = 0.01$ & 30.541 & 0.962 & 0.045 \\
$\omega = 0.1$ (Ours) & \cellcolor[HTML]{E2EFDA}\textbf{31.704} & \cellcolor[HTML]{E2EFDA}\textbf{0.967} &\cellcolor[HTML]{E2EFDA}\textbf{0.039} \\
$\omega = 1.0$  & \cellcolor[HTML]{FFF2CC}31.285 & \cellcolor[HTML]{FFF2CC}0.965 & \cellcolor[HTML]{FFF2CC}0.042 \\
\bottomrule
\end{tabular}
\caption{\textbf{Ablation on Influence Factor $\omega$}. We analyze the sensitivity of the learnable Bézier correction on the Novel-Pose Synthesis task. A value of $\omega=0.1$ provides the best balance between effective motion correction and stability. \colorbox[HTML]{E2EFDA}{Green} indicates the best and \colorbox[HTML]{FFF2CC}{yellow} indicates the second-best performance.}
\vspace{-2em}
\label{tab:ablation_alpha}
\end{table}

%% file: tables/app_results.tex
\begin{table*}[t]
\centering
\setlength{\tabcolsep}{3.5pt} 
\resizebox{\textwidth}{!}{%
\begin{tabular}{lcccc lccc}
\toprule
\multicolumn{4}{c}{\textbf{Novel-View Synthesis}} & \phantom{space} & \multicolumn{4}{c}{\textbf{Novel-Pose Synthesis}} \\
\cmidrule(r){1-4} \cmidrule(l){6-9}
\textbf{Method} & PSNR$\uparrow$ & SSIM$\uparrow$ & LPIPS$\downarrow$ && \textbf{Method} & PSNR$\uparrow$ & SSIM$\uparrow$ & LPIPS$\downarrow$ \\
\midrule

\multicolumn{9}{c}{\cellcolor[HTML]{EFEFEF}\textbf{Universal Robots UR5e}} \\ 
\midrule
2DGS~\citep{huang20242d} & 29.972 & 0.939 & 0.074 && 3DGS + FK & 26.086 & 0.945 & 0.056 \\
Robo-GS~\citep{lou2025robo} & 29.811 & 0.938 & 0.074 && 4DGS~\citep{wu20244d} & 28.239 & 0.967 & 0.039 \\
SuGaR~\citep{guedon2024sugar} & \cellcolor[HTML]{FFF2CC}30.452 & \cellcolor[HTML]{FFF2CC}0.951 & \cellcolor[HTML]{FFF2CC}0.058 && Deformable 3DGS~\citep{yang2024deformable} & \cellcolor[HTML]{FFF2CC}30.959 & \cellcolor[HTML]{FFF2CC}0.982 & \cellcolor[HTML]{FFF2CC}0.022 \\
\textbf{Ours} & \cellcolor[HTML]{E2EFDA}\textbf{35.220} & \cellcolor[HTML]{E2EFDA}\textbf{0.969} & \cellcolor[HTML]{E2EFDA}\textbf{0.025} && \textbf{Ours} & \cellcolor[HTML]{E2EFDA}\textbf{39.559} & \cellcolor[HTML]{E2EFDA}\textbf{0.989} & \cellcolor[HTML]{E2EFDA}\textbf{0.011} \\

\midrule

\multicolumn{9}{c}{\cellcolor[HTML]{EFEFEF}\textbf{ABB IRB 120}} \\
\midrule
2DGS~\citep{huang20242d} & 30.179 & \cellcolor[HTML]{FFF2CC}0.954 & 0.065 && 3DGS + FK & 24.801 & 0.933 & 0.055 \\
Robo-GS~\citep{lou2025robo} & \cellcolor[HTML]{FFF2CC}30.187 & \cellcolor[HTML]{FFF2CC}0.954 & \cellcolor[HTML]{FFF2CC}0.061 && 4DGS~\citep{wu20244d} & 30.190 & 0.966 & 0.047 \\
SuGaR~\citep{guedon2024sugar} & 28.368 & 0.950 & 0.072 && Deformable 3DGS~\citep{yang2024deformable} & \cellcolor[HTML]{FFF2CC}34.706 & \cellcolor[HTML]{FFF2CC}0.984 & \cellcolor[HTML]{FFF2CC}0.025 \\
\textbf{Ours} & \cellcolor[HTML]{E2EFDA}\textbf{34.815} & \cellcolor[HTML]{E2EFDA}\textbf{0.968} & \cellcolor[HTML]{E2EFDA}\textbf{0.038} && \textbf{Ours} & \cellcolor[HTML]{E2EFDA}\textbf{36.379} & \cellcolor[HTML]{E2EFDA}\textbf{0.982} & \cellcolor[HTML]{E2EFDA}\textbf{0.018} \\

\bottomrule
\end{tabular}%
}
\caption{\textbf{Quantitative comparison on UR5e and ABB} We evaluate our method against baselines on both Novel-View Synthesis (left) and Novel-Pose Synthesis (right). The top section shows results for the \textbf{UR5e} arm, and the bottom section for the \textbf{ABB} arm.}
\label{tab:ur5e_abb_comparison}
\end{table*}